\documentclass[a4paper]{article}

\usepackage{INTERSPEECH2022}

\usepackage[linesnumbered,ruled,vlined]{algorithm2e}
\usepackage{microtype}
\usepackage{url}

\usepackage{hyperref}

\title{Deep Sparse Conformer for Speech Recognition\thanks{Accepted by InterSpeech 2022.}}
\name{Xianchao Wu}
\address{NVIDIA}
\email{xianchaow@nvidia.com}

\begin{document}

\maketitle
\begin{abstract}
  %conformer with sparse attention and deep network with 1000 layers.
  %informer + 1000 layer transformer.
  Conformer has achieved impressive results in Automatic Speech Recognition (ASR) by leveraging transformer's capturing of content-based global interactions and convolutional neural network's exploiting of local features. In Conformer, two macaron-like feed-forward layers with half-step residual connections sandwich the multi-head self-attention and convolution modules followed by a post layer normalization. We improve Conformer's long-sequence representation ability in two directions, \emph{sparser} and \emph{deeper}. We adapt a sparse self-attention mechanism with $\mathcal{O}(L\text{log}L)$ in time complexity and memory usage. A deep normalization strategy is utilized when performing residual connections to ensure our training of hundred-level Conformer blocks. On the Japanese CSJ-500h dataset, this deep sparse Conformer achieves respectively CERs of 5.52\%, 4.03\% and 4.50\% on the three evaluation sets and 4.16\%, 2.84\% and 3.20\% when ensembling five deep sparse Conformer variants from 12 to 16, 17, 50, and finally 100 encoder layers.

  %For your paper to be published in the conference proceedings, you must use this document as both an instruction set and as a template into which you can type your own text. If your paper does not conform to the required format, you will be asked to fix it.

  %Please do not reuse your past papers as a template. To prepare your paper for submission, please always download a fresh copy of this template from the conference website and please read the format instructions in this template before you use it for your paper.

  %Conversion to PDF may cause problems in the resulting PDF or expose problems in your source document. Before submitting your final paper in PDF, check that the format in your paper PDF conforms to this template. Specifically, check the appearance of the title and author block, the appearance of section headings, document margins, column width, column spacing, and other features such as figure numbers, table numbers and equation number. In summary, you must proofread your final paper in PDF before submission.
  
  %The maximum number of pages is 5. The 5\textsuperscript{th} page is reserved for references, exclusively. However, the references may begin on an earlier page immediately after the Acknowledgements section, and continue onto the 5\textsuperscript{th} page. If no space is available on an earlier page, then the references may begin on the 5\textsuperscript{th} page.

  %Index terms should be included as shown below.
\end{abstract}

\section{Introduction}

End-to-end automatic speech recognition (ASR) systems leveraged on Transformers and their variants have achieved impressively low word/character error rates (WERs/CERs) in numerous languages during recent years \cite{Miao2020TransformerBasedOC, conformer_gulati20_interspeech, DBLP:journals/corr/abs-2010-11395, Dong2018SpeechTransformerAN,Pham2019VeryDS}. The Transformer architecture includes multi-head self-attention (MHSA) and cross-attention layers that can represent long-distance interactions inside and between sound-text sequences. Alternatively, convolutions with predefined kernel sizes capturing local contexts have also been successfully utilized in ASR \cite{Majumdar2021CitrinetCT}. 

Conformer \cite{conformer_gulati20_interspeech}, combines multi-head attentions and convolution modules, has achieved state-of-the-art accuracy in a list of benchmark datasets such as LibriSpeech. One Conformer encoder block comprises two macaron-like feed-forward network (FFN) layers with half-step residual connections. Inside these two FFNs, there is one MHSA module and one convolution module which are respectively designed for capturing global and local context information of input sound sequences. Layer normalization is further applied right after each residual connection. In the original Conformer S, M, and L models, there are 16, 16, and 17 Encoder blocks and with 10.3M, 30.7M and 118.8M parameters, respectively.

Our work is motivated by aiming at answering two questions: will \textit{deeper} Conformers achieve better accuracy, and how can we train and inference them in \textit{efficient} ways? For a sequence with length $L$, MHSA module requires to compute the ``similarity'' between every two (subsampled) ``frames'' yielding a time and memory-usage complexity of $\mathcal{O}(L^2)$. One intuitive way is to reduce the necessity of computing every pair and only compute a relatively small scale of ranges for each place in an input sequence. 

We follow this \emph{sparse attention} \cite{informer-DBLP:conf/aaai/ZhouZPZLXZ21, Li2021EfficientCS_linear_conformer, wang21ha_interspeech_probsparse} direction and use a probability attention which defines a query importance measure in MHSA and then picks only the top-$u$ (=$\mathcal{O}(\text{log}L)$) query vectors for inner-product similarity computing with key vectors. For the other \emph{deeper Conformer} direction, we borrow the \emph{deep normalization} idea used for training 1,000-layer Transformers \cite{deepnet-wang2022deepnet}. That is, the input tensor is weighted by an additional $\alpha > 1$ factor during residual connection to control and bound the model's updating for convergence. Also, another $\beta$ factor is taken as the gains for initializing the parameters in MHSA using the Xavier initialization. We name this \emph{deep sparse Conformer} and train 50-layer and 100-layer model variants under the Japanese CSJ-500h data. We report their convergence speeds during training, inference time and CERs.

%This template can be found on the conference website. Templates are provided for Microsoft Word\textregistered, and \LaTeX. However, we highly recommend using \LaTeX when preparing your submission. Information for full paper submission is available on the conference website.

\section{Related Work}\label{sec:related_work}

The \emph{ProbSparse} self-attention mechanism \cite{informer-DBLP:conf/aaai/ZhouZPZLXZ21} was first used in a Conformer-transducer for autoregressive end-to-end ASR \cite{wang21ha_interspeech}. LSTM layers were used to encode each textual sequence and a multi-layer linear module was used to combine latent representations of audios and labels. Time/memory costs' relative decrease rate reached 20\% to 50\% for long sequences from 20s to 180s. We follow their usage of \emph{ProbSparse}. The differences are that, (1) we select a non-autoregressive architecture with transformer decoders, (2) we use CTC loss and attention losses for training and CTC+attention-rescoring for decoding, and (3) we use relative positional encoding for the Conformer encoder.

A low-rank transformer was proposed in \cite{DBLP:journals/corr/abs-1910-13923} for lightweight and efficient end-to-end ASR. The linear layers used in MHSA and FFN are replaced by linear encoder-decoder units where a weight matrix $\textbf{W}\in \mathbb{R}^{m\times n}$ is approximated by two smaller matrices $\textbf{E}\in \mathbb{R}^{m\times r}$, $\textbf{D}\in \mathbb{R}^{r\times n}$ such that $\textbf{W} 
\approx \textbf{E} \times \textbf{D}$ ($r << m, n$). %When setting $r << m, n$, the number of parameters and flops in $\textbf{E}$ and $\textbf{D}$ are much smaller compared to $\textbf{W}$. 
%Experiments show that relatively larger $r$ values yield lower CERs for AiShell-1 ($r=100$ of 13.09\%) and HKUST datasets ($r=100$ of 28.95\%). We differ from this work by focusing on different parts. 
%Instead of rewriting the linear layers in MHSA and FFN, we focus on employing a sparse attention matrix. % and we build our architecture on top of a strong baseline of Conformer. % which itself improves Transformer with convolutional modules and two FFN layers in one Conformer block. 
A linear self-attention mechanism \cite{9423033-linear-attention} together with a low-rank FFN \cite{DBLP:journals/corr/abs-1910-13923} were employed in \cite{Li2021EfficientCS_linear_conformer} to build a linear-attention Conformer for ASR. The half-size models achieved comparable results with Conformer.

Compressed structures and speech attribute augmentations were used in \cite{Li2019ImprovingTS} for improving Transformer-based ASR. Parameters were shared among layers of encoder and decoder blocks. Speaker information (gender, age, education-level, speaker ID) and speech utterance properties (duration of short and long, topic of the lecture) were used to augment the training data. Using the Japanese CSJ-500h dataset, they achieved CERs of 7.6\%, 6.1\% and 6.3\% on the three test sets. We compare our models with this work since it uses CSJ dataset as well. The differences are that we do not use parameter sharing or additional speech information. %We ``compress'' transformer-style architecture in different ways.

An online compressive Transformer was utilized in \cite{leong21_interspeech} for end-to-end ASR. The input speech signal is separated into a number of blocks where the former block's compressive memory is concatenated to current block's encoder memory for decoding. Losses of CTC, RNN-transducer and attention reconstruction using transformers were minimized. %Using Aishell-1, CER of 8.7\% was achieved which is comparable with Transformer's 8.6\% on the test set.

Adaptive span self-attention was used in \cite{Chang2020EndtoEndAW} for end-to-end ASR. The motivation is to learn the appropriate span size at each self-attention head and layer during training. The limitation of spans is applied to the number of keys to be attached to each query. %With this adaptive-span applied to normal Transformer architecture, a CER of 6.7\% was achieved on Aishell's test set. %The difference is that our ``adaptiveness'' is applied to ranking the queries with a query sparseness measurement.

On the other hand, very deep models with up to 48 Transformer layers were used in \cite{Pham2019VeryDS} for end-to-end ASR. Stochastic residual connections fundamentally apply a mask $M$ on the MHSA or FFN function $F$: $R(x) = \text{LayerNorm}(M * F(x) + x)$ and $M$ only takes discrete 0 or 1 values, generated from a Bernoulli distribution similar to dropout. The difference is that we apply a factor $\alpha$ to $x$ and $\alpha$ is a function of the depth of the architecture. That is, we do not explicitly skip a MHSA or a FFN layer. %In \cite{Pham2019VeryDS}, using same 12 decoding layers, a model with 48 encoding layers (10.7\%) performed relatively worse than that with 36 encoding layers (10.4\%) on Hub5'00 test set with 300h SWB training set. 
An ensemble of n-best outputs from 36, 48 and 60 encoding layers yielded the best result of 9.9\% on the Switchboard test set.

Transformer-XL \cite{transformer-xl-dai-etal-2019-transformer} with as many as 48 layers (model size = 185.7M) was adapted in \cite{lu20g_interspeech2020_large_scale} for large-scale ASR. As the authors mentioned, the gains were not as large as they had expected and regularizing the deep transformers would be one direction for further improvements.

There are many more papers we are not able to mention here. A survey of transformer variants and their applications to end-to-end ASR can be found in \cite{DBLP:journals/corr/abs-2106-04554,DBLP:journals/corr/abs-2009-06732}. 

\section{Deeper and Sparser Conformer Blocks}

The original Conformer \cite{conformer_gulati20_interspeech} block contains two Feed Forward modules sandwiching the MHSA module and the Convolution module. This sandwich architecture is inspired by Macaron-Net \cite{macaron-net-lu2019understanding}, in which the original feed-forward layer in one Transformer layer is separated into two half-step feed-forward layers, one before the self-attention layer and another after. Mathematically, for an input $x_i$ sent to a Conformer block $i$, the output $y_i$ is:
\begin{align}
    \tilde{x}_i & = x_i + 0.5\times \text{FFN}(x_i) \\
    {x'_i} & = \tilde{x}_i + \text{MHSA}(\tilde{x}_i) \label{eq_mhsa_orig_conformer}\\
    {x''_i} & = x'_i + \text{Conv}(x'_i) \\
    {y_i} & = \text{LayerNorm}(x''_i + 0.5\times \text{FFN}(x''_i)) 
\end{align}

There are four residual connections in one Conformer block and a layer normalization (LN) is used finally (named as Post-LN). As has been observed and reported in \cite{deepnet-wang2022deepnet, xiong-icml2020-DBLP:conf/icml/XiongYHZZXZLWL20}, this Post-LN has a problem of unstable training. Furthermore, the MHSA network requires a time and memory-usage complexity of $\mathcal{O}(L^2)$ which is barely acceptable when we have relatively long sequences in our training data or during inferencing.

\subsection{Sparser Attention}

There are a list of work that proposed new attention mechanisms to replace the $\mathcal{O}(L^2)$ time/space complexities into $\mathcal{O}(L\text{log}L)$ or even $\mathcal{O}(L)$ \cite{Li2021EfficientCS_linear_conformer, wang21ha_interspeech_probsparse}. Motivated by \cite{informer-DBLP:conf/aaai/ZhouZPZLXZ21} for modeling long sequences for time-series forecasting, we adapt the \emph{ProbSparse} self-attention mechanism to replace the MHSA function in Equation \ref{eq_mhsa_orig_conformer}. In \emph{ProbSparse}, the original tensor $\textbf{Q}\in \mathbb{R}^{b\times L \times d}$ ($b$ is batch size, $L$ is input sequence length, and $d$ is hidden dimension. For simplicity, we set $b=1$ and omit $b$ hereafter.) is replaced by a same shape tensor $\bar{\textbf{Q}}$ which only contains top-$u$ ($=\mathcal{O}(\text{log}L)$) queries under a sparsity measurement $M(\textbf{q}, \textbf{K})$. That is, each key is only allowed to attend to $u$ dominant queries:
\begin{align}
    \mathcal{A}(\textbf{Q}, \textbf{K}, \textbf{V}) = & \sum_j \Big[ \frac{k(\textbf{Q}, \textbf{k}_j)}{\sum_lk(\textbf{Q}, \textbf{k}_l)} \Big] \textbf{v}_j\\ 
    = &\ \mathbb{E}_{p(\textbf{k}_j|\textbf{Q})}[\textbf{v}_j]\\
    \approx &\ \text{Softmax}(\frac{\bar{\textbf{Q}}\textbf{K}^\top}{\sqrt{d}})\textbf{V} \label{eq_softmax_Q_bar}.
\end{align}
Here, $\textbf{q}_i, \textbf{k}_j\in \mathbb{R}^{1\times h}$ are row vectors, and $k(\textbf{q}_i \in \bar{\textbf{Q}}, \textbf{k}_j)$ is selected to be an exponential kernel $\text{exp}(\textbf{q}_i\textbf{k}_j^\top/\sqrt{d})$.

The query sparsity measure $M(\textbf{q}, \textbf{K})$ is motivated by (1) the dominant dot-product pairs encouraging the corresponding query's attention probability distribution $p(\textbf{k}_j|\textbf{q}_i)$ \emph{away from} the uniform distribution, and (2) Kullback-Leibler divergence $KL(q||p)$ measuring the ``distance'' between $p(\textbf{k}_j|\textbf{q}_i)$ and uniform distribution $q(\textbf{k}_j|\textbf{q}_i)=1/L$. An empirical approximation for efficiently computing the query sparsity measurement for $\textbf{q}_i$ is defined as the following ``max-mean'' equation:
\begin{equation}
    \bar{M}{(\textbf{q}_i, \textbf{K})}=\text{max}_j\{\frac{{\textbf{q}_i}\textbf{k}_j^\top}{\sqrt{d}}\} - \frac{1}{L}\sum_{j=1}^{L}\frac{{\textbf{q}_i}\textbf{k}_j^\top}{\sqrt{d}}. \label{eq_max_mean_Q}
\end{equation}

The range of top-$u$ approximately holds as proved in \cite{informer-DBLP:conf/aaai/ZhouZPZLXZ21}. Under the long tail distribution, it has been empirically testified that randomly sample $U=L\text{log}L$ dot-product pairs to compute the query sparsity measure and then select sparse top-$u$ from it and form $\bar{\textbf{Q}}$ yielded acceptable results \cite{wang21ha_interspeech_probsparse}.

We use the sparse attention function (MHSA-Sparse) defined by Equation \ref{eq_softmax_Q_bar} and assisted by Equation \ref{eq_max_mean_Q} to replace the original MHSA function used in Equation \ref{eq_mhsa_orig_conformer}:
\begin{align}
    \text{MHSA-Sparse}(\textbf{X}) & = \text{Concat}(\text{head}_1, ..., \text{head}_h)\textbf{W}^\textbf{O} \\
    \text{where,}\ \text{head}_{i\in\{1,...,h\}} &=\mathcal{A}(\textbf{X}\textbf{W}_i^\textbf{Q}, \textbf{X}\textbf{W}_i^\textbf{K}, \textbf{X}\textbf{W}_i^\textbf{V})
\end{align}
Here, the linear projections are trainable parameter matrices $\textbf{W}_i^\textbf{Q}$ and $\textbf{W}_i^\textbf{K}\in \mathbb{R}^{d\times d_k}$, $\textbf{W}_i^\textbf{V}\in \mathbb{R}^{d\times d_v}$, $\textbf{W}^\textbf{O}\in \mathbb{R}^{hd_v\times d}$ and $d_k=d_v=d/h$.

Algorithm \ref{alg:probattn_relpos_attention} describes the pseudo-code of a multi-head attention layer using \emph{ProbSparse} (lines 12 to 19, 21) and relative position encoding (lines 10, 11, 19). One additional linear layer with weight matrix $\textbf{W}^\textbf{P}\in \mathbb{R}^{d\times d}$ is introduced to parameterize the memory tensors' position information. Following \cite{transformer-xl-dai-etal-2019-transformer}, two learnable vectors $\textbf{U}_1$, $\textbf{U}_2 \in \mathbb{R}^d$ are used in shapes of $(h, d_q)$ in line 10 and 11. They are introduced to assist relative position encoding by alleviating attentive bias towards different words in different positions. 

Equation \ref{eq_max_mean_Q} suggests an inner-product computation of between a query vector to all the key vectors to determine the importance of the query vector. This blocks the reduction of time complexity. Instead, we only use $L'_{\textbf{K}}$ number of key vectors through sampling from the whole $L_\textbf{K}$ (line 12, 13). This ensures a time complexity of $\mathcal{O}(L_\textbf{Q}\text{ln}L_\textbf{K})$ for computing Equation \ref{eq_max_mean_Q} in line 15. A hyper-parameter $c_1$ (= 5.0) determines the size of the key vector subset.

When the query sparseness measurement is computed, we can easily pick the top-$L'_{\textbf{Q}}$ queries where $L'_{\textbf{Q}}$ is the number of query vectors selected. A hyper-parameter $c_2$ (= 5.0) determines the size of the query vector subset. Line 19 combines the ``similarity scores'' between the reduced query vector set and the original key and position tensors. Then, the scores are masked by the lengths of sequences in a batch, normalized into probability-style scores under Softmax and finally dropped out with a given probability $p_{dropout}$.

\begin{algorithm}[t!]  
	\caption{ProbSparse+Relative Position Encode}  
	\label{alg:probattn_relpos_attention}
	Given: \textbf{X}, $\textbf{X}_{\textbf{P}}$, \textbf{Y}, $c_1$, $c_2$, $mask$, $p_{dropout}$\\
	Trainable parameters: $\textbf{W}^{\{\textbf{Q}, \textbf{K}, \textbf{V}, \textbf{P}, \textbf{O}\}}$, $\textbf{U}_{\{1,2\}}$\\
	\textbf{Q}, \textbf{K}, \textbf{V}, \textbf{P} = \textbf{Y}$\textbf{W}^\textbf{Q}$, \textbf{X}$\textbf{W}^\textbf{K}$, \textbf{X}$\textbf{W}^\textbf{V}$, $\textbf{X}_\textbf{p} \textbf{W}^\textbf{P}$  $\triangleright$ $\textbf{X}_\textbf{P}\in \mathbb{R}^{b \times L_\textbf{K} \times d}$ is the relative position encoding tensor for \textbf{X}, and \textbf{Y}=\textbf{X} for self-attention \\
	$b$, $L_\textbf{Q}$, $h\times d_q$ = \textbf{Q}.shape \\
	$L_\textbf{K}$ = \textbf{K}.shape[1] \\
	\textbf{Q} = \textbf{Q}.view($b$, $L_\textbf{Q}$, $h$, $d_q$) \\
	\textbf{K} = \textbf{K}.view($b$, $L_\textbf{K}$, $h$, $d_k$).transpose(1,2) \\
	\textbf{V} = \textbf{V}.view($b$, $L_\textbf{K}$, $h$, $d_v$).transpose(1,2) \\
	\textbf{P} = \textbf{P}.view($b$, $L_\textbf{K}$, $h$, $d_k$).transpose(1,2) \\
	$\textbf{Q}_{\textbf{U}_1}$ = (\textbf{Q} + $\textbf{U}_1$).transpose(1,2) $\triangleright$ $\textbf{U}_1\in \mathbb{R}^{h\times d_q}$ \\
	
	$\textbf{Q}_{\textbf{U}_2}$ = (\textbf{Q} + $\textbf{U}_2$).transpose(1,2) $\triangleright$ $\textbf{U}_2\in \mathbb{R}^{h\times d_q}$ \\
	
	${L'}_\textbf{K}$ = $c_1 \times \lceil \ln(L_\textbf{K}) \rceil$ \\
	
	$\textbf{K}_{part}$ = Sample($\textbf{K}$, ${L'}_\textbf{K}$, dim=-2) $\triangleright$ $\textbf{K}_{part} \in \mathbb{R}^{b\times h \times {L'}_\textbf{K} \times d_k}$ \\
	${L'}_\textbf{Q}$ = $c_2 \times \lceil \ln(L_\textbf{Q}) \rceil$ \\
	
	$\bar{\textbf{M}}$ = ($\textbf{Q}_{\textbf{U}_1}\textbf{K}^\top_{part}$).max(dim=-1)[0] - ($\textbf{Q}_{\textbf{U}_1}\textbf{K}^\top_{part}$).sum(-1)/$L_\textbf{K}$ $\triangleright$ Equation \ref{eq_max_mean_Q}, omit $\sqrt{d}$ \\
	$\bar{\textbf{M}}_{top}$ = $\bar{\textbf{M}}$.topk(${L'}_\textbf{Q}$)[1] $\triangleright$ index of top-${L'}_\textbf{Q}$ queries \\
	
	$\bar{\textbf{Q}}_{\textbf{U}_1}$ = $\textbf{Q}_{\textbf{U}_1}$[$\bar{\textbf{M}}_{top}$, :] $\triangleright$ $\bar{\textbf{Q}}_{\textbf{U}_1}\in \mathbb{R}^{b\times h\times {L'}_\textbf{Q}\times d_q}$ \\
	
	$\bar{\textbf{Q}}_{\textbf{U}_2}$ = $\textbf{Q}_{\textbf{U}_2}$[$\bar{\textbf{M}}_{top}$, :] \\

	\textbf{S} = ($\bar{\textbf{Q}}_{\textbf{U}_1} \textbf{K}^\top$ + $\bar{\textbf{Q}}_{\textbf{U}_2}\textbf{P}^\top$) / $\sqrt{d_q}$ $\triangleright$ $\textbf{S} \in \mathbb{R}^{b\times h \times {L'}_\textbf{Q} \times L_\textbf{K}}$ \\
	\textbf{S} = Dropout(Softmax(Mask(\textbf{S}, $mask$)), $p_{dropout}$) \\
	\textbf{V}[$\bar{\textbf{M}}_{top}$, :] = \textbf{S}\textbf{V} $\triangleright$ only change a part of \textbf{V} \\
	
	\textbf{V} = \textbf{V}.transpose(1,2).contiguous().view($b$, -1, $h\times d_v$) \\
	\textbf{V} = \textbf{V}$\textbf{W}^\textbf{O}$ \\
	return \textbf{V} \\

\end{algorithm}

\subsection{Deeper Normalization}

The Conformer-Large defined in \cite{conformer_gulati20_interspeech} contains 118.8M parameters in which there are 17 encoder layers with encoder dimension of 512, 1 decoder layer with decoder dimension of 640. Deeper models alike (1) GPT-3 \cite{gpt3_NEURIPS2020_1457c0d6} model with 175B parameters and 96 transformer layers has achieved impressive scores both for classification and generation tasks through few-shot prompt-based learning, (2) DeepNet \cite{deepnet-wang2022deepnet} with 1,000 transformer layers has shown significantly better BLEU scores \cite{papineni-etal-2002-bleu} in multilingual neural machine translation, compared with shallow architectures. Inspired by these, we aim at enlarging the number of layers in Conformer to as much as one-hundred levels and testifying its representation ability for ASR.

DeepNet \cite{deepnet-wang2022deepnet} combines the good performance of Post-LN and stable training of Pre-LN by leveraging a new normalization function named DeepNorm when performing residual connections \cite{resnet-He2016DeepRL} in a Transformer layer:
\begin{equation}
    \text{DeepNorm}(x) = \text{LayerNorm}(x\times \alpha + f(x)).
\end{equation}
In order to ensure the model updating in DeepNet to be constrained by bounds for stable training, $\alpha$ is assigned with various values alike $(2N)^{\frac{1}{4}}$ for encoder-only framework (e.g., BERT \cite{bert-devlin-etal-2019-bert}), and $(2M)^{\frac{1}{4}}$ for decoder-only framework (e.g., GPT \cite{gpt3_NEURIPS2020_1457c0d6}) where $N$ and $M$ respectively stand for the number of $f(\cdot)$, such as self-attention, cross-attention or FFN sub-layers.

\begin{figure}[t]
  \centering
  \includegraphics[width=6.5cm]{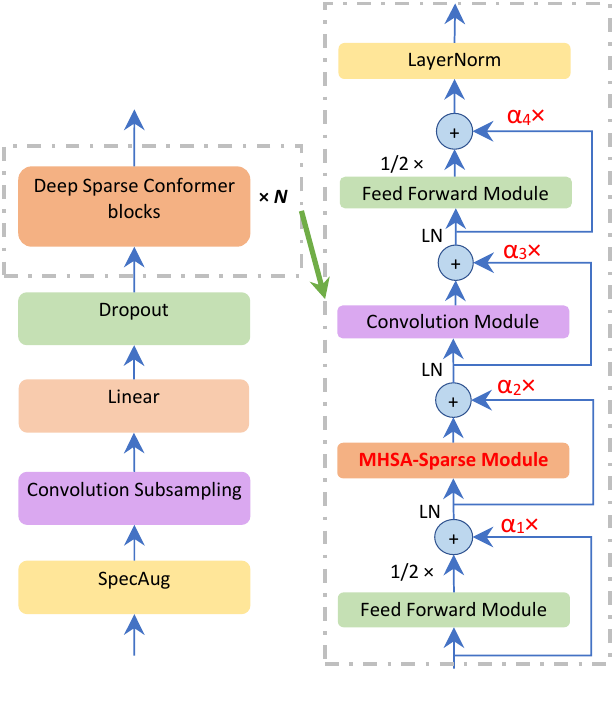}
  \caption{Encoder model architecture for our deep sparse Conformer. Updates are marked in red color: $\alpha_i$ on $x$ and using sparse attention in MHSA. }
  \label{fig:deep_sparse_conformer}
\end{figure}

We adapt this DeepNorm function in our deep sparse Conformer. The updated Conformer block is mathematically expressed as:
\begin{align}
    \tilde{x}_i & = \text{LayerNorm}(x_i\times \alpha_1  + 0.5\times \text{FFN}(x_i)) \\
    {x'_i} & = \text{LayerNorm}(\tilde{x}_i \times \alpha_2 + \text{MHSA-Sparse}(\tilde{x}_i)) \label{eq_mhsa_new_conformer}\\
    {x''_i} & = \text{LayerNorm}(x'_i\times \alpha_3  + \text{Conv}(x'_i)) \\
    {y_i} & = \text{LayerNorm}(x''_i\times \alpha_4 + 0.5\times \text{FFN}(x''_i))
\end{align}
Here, we utilize $\alpha_{1,2,3,4}$, all take a value of $0.81(N^4M)^{1/16}$ ($N$/$M$=encoder/decoder layer numbers), to scale the temporary output from former sub-layer for residual connection. MHSA-Sparse function is defined in Equation \ref{eq_softmax_Q_bar} and described in Algorithm \ref{alg:probattn_relpos_attention}. %In addition, we change Conformer's usage of one Pre-LN to Post-LN in each module of one block. 
Figure \ref{fig:deep_sparse_conformer} depicts the encoder model architecture of our deep sparse Conformer in which one Conformer block is extended into a pipeline of four modules with four residual connections controlled by DeepNorm.

\section{Experiments}

\subsection{Setup}

Our experiments are performed under an open-source ASR platform, WeNet\footnote{\url{https://github.com/wenet-e2e/wenet}} \cite{wenet-DBLP:journals/corr/abs-2102-01547}. We release code and pretrained models\footnote{\url{https://github.com/Xianchao-Wu/wenet-deep-sparse-conformer}}. We select a Conformer baseline model consists of a 12-block Conformer encoder ($d_{\text{FFN}}$=2048, $h$=8, $d$=512, CNN$_{\text{kernel}}$=31) and a 3-block bidirectional Transformer decoder ($d_{\text{FFN}}$=2048, $h$=8, $d$=512) which encodes the textual sequences in left-to-right ($l2r$) and right-to-left ($r2l$) directions. The objective is to minimize $\mathcal{L}$, a linear combination of the CTC loss \cite{ctc-10.1145/1143844.1143891} ($\lambda$=0.3) and attention losses (ATT) computed by point-wise KL-divergence \cite{KL-10.1214/aoms/1177729694}:
\begin{equation}
    \mathcal{L} = 0.3*\text{CTC}_{loss} + 0.7*(0.7*\text{ATT}_{l2r} + 0.3*\text{ATT}_{r2l})
\end{equation}
Label smoothing with $\delta=0.1$ is applied to the attention objective so that the references are discounted by (1-$\delta$).

During training, we select the Adam optimizer \cite{Kingma2015AdamAM} with a maximum learning rate of 0.002. The Noam learning rate scheduler with 30K warm-up steps is used. The models are trained with static batching skill for 120 epochs. The top 30 best checkpoints ranked by validation set losses are averaged to be the final model. For decoding, we use CTC greedy search and Attention rescoring. All our experiments were performed under NVIDIA DGX-A100 with 8*A100-80GB GPUs.
%\footnote{\url{https://www.nvidia.com/en-us/data-center/dgx-a100/}}

\subsection{Data}

We evaluate deep sparse Conformer variants on the Japanese ``Corpus of Spontaneous Japanese'' (CSJ) dataset\footnote{\url{https://ccd.ninjal.ac.jp/csj/en/}}, which consists of 500 hours of labeled speech. We use sentencepiece-bpe \cite{kudo-richardson-2018-sentencepiece} to segment the text sequences and select a vocabulary size of 4,096. During data preparation, we generate 80-dimension FBank feature vectors with a 25ms window, a 10ms frame stride and dither=1.0. SpecAugment \cite{spec-aug-2019} is adapted with 2 frequency masks ($F$=10) and 2 time masks ($T$=50). Global Cepstral Mean and Variance Normalization \cite{cmvn-10.1016/S0167-6393(98)00033-8} technique is employed to normalize the 80-dimension FBank feature vectors.

\subsection{Major Results on CSJ Tests}

\begin{table}%[th]
\footnotesize
  \caption{CERs (\%) of 4 baselines and 5 deep sparse Conformer (DSC) variants trained with the CSJ-500h dataset.}
  \label{tab:main_res_compare}
  \centering
  \begin{tabular}{ccccccl}
 \toprule
 & 1.73h & 1.82h & 1.23h & train. & inf. & size\\ 
models & Test1 & Test2 & Test3 & /ep. & char/s & (M)\\ 
%\midrule
\hline\hline
Trans.\cite{Li2019ImprovingTS} & 7.6 & 6.1 & 6.3 & - & - & 36\\ 
Espnet+LM & 6.5 & 4.6 & 5.1 & - & - & -\\ 
Citrinet\cite{Majumdar2021CitrinetCT} & 7.28 & 4.81 & 5.44 & 5m & 210 & 22.7\\ 
Conformer\cite{conformer_gulati20_interspeech} & 6.97 & 4.65 & 5.29 & 10m & 194 & 135.1\\ 
\midrule
DSC-small (16) & 7.64 & 5.35 & 6.15 & 6m & 310 & 14.2\\ 
DSC-12en & 5.52 & 4.03 & 4.50 & 8m & 239 & 135.2\\ 
DSC-17en & 6.00 & 4.30 & 5.17 & 13m & 235 & 169.4\\ 
DSC-50en & 6.20 & 4.31 & 5.49 & 42m & 203 & 395.4\\ 
DSC-100en & 6.27 & 4.36 & 5.56 & 132m & 180 & 737.8\\ 
Ensemble+LM & \textbf{4.16} & \textbf{2.84} & \textbf{3.20} & - & - & -\\ 
\bottomrule
  \end{tabular}
\end{table}

Table \ref{tab:main_res_compare} lists the CERs of five of our deep sparse Conformer (DSC) variants and four baselines. The first baseline is a Transformer with enhanced information from speakers and speech \cite{Li2019ImprovingTS} (refer to Section \ref{sec:related_work}). The second baseline is Espnet's implementation of deep VGGBLSTM with CTC joint decoding and LM rescoring\footnote{\url{https://github.com/espnet/espnet/blob/master/egs/csj/asr1/RESULTS.md}}. Note that only this baseline uses an external language model among all the nine models. The third baseline is Citrinet \cite{Majumdar2021CitrinetCT} (channel size = 384), a recent pure CNN-based model with 1D time-channel separable convolutions combined with squeeze-and-excitation and has achieved comparable results compared with Conformer. The fourth baseline is Conformer-Large \cite{conformer_gulati20_interspeech} with a bi-decoder. 

Our first DSC variant is a Conformer-Small model with the same configurations of Conformer-S \cite{conformer_gulati20_interspeech} with 16 layers except that we use a bidirectional decoder. The other four variants differ only at the encoder part of different layers: 12, 17, 50, and 100. These 5 variants use a bidirectional decoder in which both $l2r$ and $r2l$ decoders have 3 Transformer decoder layers. 

We have the following observations. First, Conformer performs well among the baselines, with a relatively large model size. Second, considering that both Conformer and DSC-12en have 135M parameter sizes, we can compare their CERs and training/inferencing time directly. DSC-12en achieved an absolute CER improvement of 1.45\%, 0.62\% and 0.79\% on the three test sets. Also, the training time is only 80\% of Conformer's and the inference speed is 1.23 times faster. These improvements of accuracy and training/inferencing speed reflect the effectiveness of the deep sparse Conformer blocks. Third, using DeepNorm strategy, we could successfully train 50 and 100 layers of Conformer. However, the training time increases significantly: the training time for DSC-17en, DSC-50en and DSC-100en are respectively 1.6, 5.25 and 16.5 times of DSC-12en's. The accuracies slightly dropped when we enlarged the model parameter sizes. This observation aligns with that reported in \cite{Pham2019VeryDS}. Following \cite{Pham2019VeryDS}, we also performed 6-gram LM enhanced ensembles of 30-best outputs from the 5 variants and achieved the best CERs for the 3 test sets: 4.16\%, 2.84\% and 3.20\%. 

%report their results here.

\subsection{Ablation Studies}

%\subsubsection{Conformer Block vs. Deep Sparse Conformer Block}

%The differences include, sparse attention, deep norm, show their performances of speed and accuracies.

\subsubsection{Number of Deep Sparse Conformer Blocks}

\begin{table}%[th]
\footnotesize
  \caption{System ensemble CERs (\%) with two types of decoding: attention rescoring (Attn) and CTC greedy searching.}
  \label{tab:ensemble_res_compare}
  \centering
  \begin{tabular}{ccccccc}
\toprule
 & \multicolumn{2}{c}{Test1}  & \multicolumn{2}{c}{Test2}  & \multicolumn{2}{c}{Test3} \\ 
 & Attn & CTC & Attn & CTC & Attn & CTC \\ 
 \midrule
small-1 & 7.64 & 7.82 & 5.35 & 5.53 & 6.15 & 6.40\\ 
12en-2 & 5.52 & 5.62 & 4.03 & 4.23 & 4.50 & 4.39\\ 
17en-3 & 6.00 & 6.16 & 4.30 & 4.51 & 5.17 & 5.02\\ 
\midrule
1+2+3 & 4.83 & 4.94 & 3.36 & 3.51 & 3.74 & 3.70\\ 
+50en & 4.38 & 4.45 & 3.00 & 3.16 & 3.40 & 3.29\\ 
++100en & \textbf{4.16} & 4.21 & \textbf{2.84} & 2.92 & 3.20 & \textbf{3.09}\\  
\bottomrule
  \end{tabular}
\end{table}

Table \ref{tab:ensemble_res_compare} shows the detailed CER improvements when (1) ensembling DSC-small, DSC-12en and DSC-17en, (2) further appending DSC-50en and (3) finally adding DSC-100en. By ensembling the first three models, CER improved 0.70\% averagely. Then, adding DSC-50en brings a gain of 0.40\% on average. Finally, the DSC-100en can further contribute 0.21\% improvements on average. These reflect that larger models are still helpful yet their contributions are marginally decreased. %It also reflects that making a small improvement on strong baselines is relatively more difficult.

%\subsubsection{Sampling Factors in Sparse Attentions}

%show variants of $c_1$ and $c_2$ and the final result variants.

\subsubsection{Pre-LN vs. Post-LN in Deep Normalization}

We investigated Pre-LN and Post-LN when performing DeepNorm. Different from that observed in NLP fields, Post-LN was less stable in DSCs and the loss could not converge after 30 epochs. We thus select a mixture solution in our five model variants by (1) keeping the LNs in Post-LN and (2) initially adding a Pre-LN layer.

\section{Conclusions}

%Authors must proofread their PDF file prior to submission to ensure it is correct. Authors should not rely on proofreading the Word file. Please proofread the PDF file before it is submitted.

In this paper, we introduced deep sparse Conformer, an architecture that integrates (1) \emph{sparse} attentions guided by a query sparseness measure and (2) \emph{deep} normalization which weight input tensors during residual connection into Conformer. We demonstrated that the usage of sparse attentions and deep normalization yielded faster training speed and stable training of much deeper Conformer encoder blocks. We built model variants with as much as 100 encoder layers and trained them using the Japanese CSJ-500h datasets. Future work includes using deep sparse Conformer blocks in self-supervised speech pretraining and fine-tuning. 

%\section{Acknowledgements}

%The ISCA Board would like to thank the organizing committees of the past INTERSPEECH conferences for their help and for kindly providing the template files. Note to authors: Authors should not use logos in the acknowledgement section; rather authors should acknowledge corporations by naming them only.

\bibliographystyle{IEEEtran}

\bibliography{mybib}

% Generated by IEEEtran.bst, version: 1.13 (2008/09/30)
\begin{thebibliography}{10}
\providecommand{\url}[1]{#1}
\csname url@samestyle\endcsname
\providecommand{\newblock}{\relax}
\providecommand{\bibinfo}[2]{#2}
\providecommand{\BIBentrySTDinterwordspacing}{\spaceskip=0pt\relax}
\providecommand{\BIBentryALTinterwordstretchfactor}{4}
\providecommand{\BIBentryALTinterwordspacing}{\spaceskip=\fontdimen2\font plus
\BIBentryALTinterwordstretchfactor\fontdimen3\font minus
  \fontdimen4\font\relax}
\providecommand{\BIBforeignlanguage}[2]{{%
\expandafter\ifx\csname l@#1\endcsname\relax
\typeout{** WARNING: IEEEtran.bst: No hyphenation pattern has been}%
\typeout{** loaded for the language `#1'. Using the pattern for}%
\typeout{** the default language instead.}%
\else
\language=\csname l@#1\endcsname
\fi
#2}}
\providecommand{\BIBdecl}{\relax}
\BIBdecl

\bibitem{Miao2020TransformerBasedOC}
H.~Miao, G.~Cheng, C.~Gao, P.~Zhang, and Y.~Yan, ``Transformer-based online
  ctc/attention end-to-end speech recognition architecture,'' \emph{ICASSP
  2020}, pp. 6084--6088, 2020.

\bibitem{conformer_gulati20_interspeech}
A.~Gulati, J.~Qin, C.-C. Chiu, N.~Parmar, Y.~Zhang, J.~Yu, W.~Han, S.~Wang,
  Z.~Zhang, Y.~Wu, and R.~Pang, ``{Conformer: Convolution-augmented Transformer
  for Speech Recognition},'' in \emph{Proc. Interspeech 2020}, 2020, pp.
  5036--5040.

\bibitem{DBLP:journals/corr/abs-2010-11395}
\BIBentryALTinterwordspacing
X.~Chen, Y.~Wu, Z.~Wang, S.~Liu, and J.~Li, ``Developing real-time streaming
  transformer transducer for speech recognition on large-scale dataset,''
  \emph{CoRR}, vol. abs/2010.11395, 2020. [Online]. Available:
  \url{https://arxiv.org/abs/2010.11395}
\BIBentrySTDinterwordspacing

\bibitem{Dong2018SpeechTransformerAN}
L.~Dong, S.~Xu, and B.~Xu, ``Speech-transformer: A no-recurrence
  sequence-to-sequence model for speech recognition,'' \emph{2018 ICASSP}, pp.
  5884--5888, 2018.

\bibitem{Pham2019VeryDS}
N.-Q. Pham, T.~S. Nguyen, J.~Niehues, M.~M{\"u}ller, and A.~H. Waibel, ``Very
  deep self-attention networks for end-to-end speech recognition,'' in
  \emph{INTERSPEECH}, 2019.

\bibitem{Majumdar2021CitrinetCT}
S.~Majumdar, J.~Balam, O.~Hrinchuk, V.~Lavrukhin, V.~Noroozi, and B.~Ginsburg,
  ``Citrinet: Closing the gap between non-autoregressive and autoregressive
  end-to-end models for automatic speech recognition,'' 2021.

\bibitem{informer-DBLP:conf/aaai/ZhouZPZLXZ21}
\BIBentryALTinterwordspacing
H.~Zhou, S.~Zhang, J.~Peng, S.~Zhang, J.~Li, H.~Xiong, and W.~Zhang,
  ``Informer: Beyond efficient transformer for long sequence time-series
  forecasting,'' in \emph{AAAI, 2021}.\hskip 1em plus 0.5em minus 0.4em\relax
  {AAAI} Press, 2021, pp. 11\,106--11\,115. [Online]. Available:
  \url{https://ojs.aaai.org/index.php/AAAI/article/view/17325}
\BIBentrySTDinterwordspacing

\bibitem{Li2021EfficientCS_linear_conformer}
S.~Li, M.~Xu, and X.-L. Zhang, ``Efficient conformer-based speech recognition
  with linear attention,'' \emph{2021 Asia-Pacific Signal and Information
  Processing Association Annual Summit and Conference (APSIPA ASC)}, pp.
  448--453, 2021.

\bibitem{wang21ha_interspeech_probsparse}
X.~Wang, S.~Sun, L.~Xie, and L.~Ma, ``{Efficient Conformer with Prob-Sparse
  Attention Mechanism for End-to-End Speech Recognition},'' in \emph{Proc.
  Interspeech 2021}, 2021, pp. 4578--4582.

\bibitem{deepnet-wang2022deepnet}
H.~Wang, S.~Ma, L.~Dong, S.~Huang, D.~Zhang, and F.~Wei, ``Deepnet: Scaling
  transformers to 1,000 layers,'' \emph{arXiv preprint arXiv:2203.00555}, 2022.

\bibitem{DBLP:journals/corr/abs-1910-13923}
\BIBentryALTinterwordspacing
G.~I. Winata, S.~Cahyawijaya, Z.~Lin, Z.~Liu, and P.~Fung, ``Lightweight and
  efficient end-to-end speech recognition using low-rank transformer,''
  \emph{CoRR}, vol. abs/1910.13923, 2019. [Online]. Available:
  \url{http://arxiv.org/abs/1910.13923}
\BIBentrySTDinterwordspacing

\bibitem{9423033-linear-attention}
S.~Zhuoran, Z.~Mingyuan, Z.~Haiyu, Y.~Shuai, and L.~Hongsheng, ``Efficient
  attention: Attention with linear complexities,'' in \emph{2021 IEEE Winter
  Conference on Applications of Computer Vision (WACV)}, 2021, pp. 3530--3538.

\bibitem{Li2019ImprovingTS}
S.~Li, D.~Raj, X.~Lu, P.~Shen, T.~Kawahara, and H.~Kawai, ``Improving
  transformer-based speech recognition systems with compressed structure and
  speech attributes augmentation,'' in \emph{INTERSPEECH}, 2019.

\bibitem{leong21_interspeech}
C.-H. Leong, Y.-H. Huang, and J.-T. Chien, ``{Online Compressive Transformer
  for End-to-End Speech Recognition},'' in \emph{Proc. Interspeech 2021}, 2021,
  pp. 2082--2086.

\bibitem{Chang2020EndtoEndAW}
X.~Chang, A.~S. Subramanian, P.~Guo, S.~Watanabe, Y.~Fujita, and M.~Omachi,
  ``End-to-end asr with adaptive span self-attention,'' in \emph{INTERSPEECH},
  2020.

\bibitem{transformer-xl-dai-etal-2019-transformer}
\BIBentryALTinterwordspacing
Z.~Dai, Z.~Yang, Y.~Yang, J.~Carbonell, Q.~Le, and R.~Salakhutdinov,
  ``Transformer-{XL}: Attentive language models beyond a fixed-length
  context,'' in \emph{Proceedings of the 57th Annual Meeting of the Association
  for Computational Linguistics}.\hskip 1em plus 0.5em minus 0.4em\relax
  Florence, Italy: Association for Computational Linguistics, Jul. 2019, pp.
  2978--2988. [Online]. Available: \url{https://aclanthology.org/P19-1285}
\BIBentrySTDinterwordspacing

\bibitem{lu20g_interspeech2020_large_scale}
L.~Lu, C.~Liu, J.~Li, and Y.~Gong, ``{Exploring Transformers for Large-Scale
  Speech Recognition},'' in \emph{Proc. Interspeech 2020}, 2020, pp.
  5041--5045.

\bibitem{DBLP:journals/corr/abs-2106-04554}
\BIBentryALTinterwordspacing
T.~Lin, Y.~Wang, X.~Liu, and X.~Qiu, ``A survey of transformers,'' \emph{CoRR},
  vol. abs/2106.04554, 2021. [Online]. Available:
  \url{https://arxiv.org/abs/2106.04554}
\BIBentrySTDinterwordspacing

\bibitem{DBLP:journals/corr/abs-2009-06732}
\BIBentryALTinterwordspacing
Y.~Tay, M.~Dehghani, D.~Bahri, and D.~Metzler, ``Efficient transformers: {A}
  survey,'' \emph{CoRR}, vol. abs/2009.06732, 2020. [Online]. Available:
  \url{https://arxiv.org/abs/2009.06732}
\BIBentrySTDinterwordspacing

\bibitem{macaron-net-lu2019understanding}
Y.~Lu, Z.~Li, D.~He, Z.~Sun, B.~Dong, T.~Qin, L.~Wang, and T.-Y. Liu,
  ``Understanding and improving transformer from a multi-particle dynamic
  system point of view,'' \emph{arXiv preprint arXiv:1906.02762}, 2019.

\bibitem{xiong-icml2020-DBLP:conf/icml/XiongYHZZXZLWL20}
\BIBentryALTinterwordspacing
R.~Xiong, Y.~Yang, D.~He, K.~Zheng, S.~Zheng, C.~Xing, H.~Zhang, Y.~Lan,
  L.~Wang, and T.~Liu, ``On layer normalization in the transformer
  architecture,'' in \emph{Proceedings of ICML}, ser. Proceedings of Machine
  Learning Research, vol. 119.\hskip 1em plus 0.5em minus 0.4em\relax {PMLR},
  2020, pp. 10\,524--10\,533. [Online]. Available:
  \url{http://proceedings.mlr.press/v119/xiong20b.html}
\BIBentrySTDinterwordspacing

\bibitem{gpt3_NEURIPS2020_1457c0d6}
T.~Brown, B.~Mann, N.~Ryder, M.~Subbiah, J.~D. Kaplan, P.~Dhariwal,
  A.~Neelakantan, P.~Shyam, G.~Sastry, A.~Askell, S.~Agarwal, A.~Herbert-Voss,
  G.~Krueger, T.~Henighan, R.~Child, A.~Ramesh, D.~Ziegler, J.~Wu, C.~Winter,
  C.~Hesse, M.~Chen, E.~Sigler, M.~Litwin, S.~Gray, B.~Chess, J.~Clark,
  C.~Berner, S.~McCandlish, A.~Radford, I.~Sutskever, and D.~Amodei, ``Language
  models are few-shot learners,'' in \emph{Advances in Neural Information
  Processing Systems}, vol.~33.\hskip 1em plus 0.5em minus 0.4em\relax Curran
  Associates, Inc., 2020, pp. 1877--1901.

\bibitem{papineni-etal-2002-bleu}
\BIBentryALTinterwordspacing
K.~Papineni, S.~Roukos, T.~Ward, and W.-J. Zhu, ``{B}leu: a method for
  automatic evaluation of machine translation,'' in \emph{Proceedings of
  ACL}.\hskip 1em plus 0.5em minus 0.4em\relax Philadelphia, Pennsylvania, USA:
  Association for Computational Linguistics, Jul. 2002, pp. 311--318. [Online].
  Available: \url{https://aclanthology.org/P02-1040}
\BIBentrySTDinterwordspacing

\bibitem{resnet-He2016DeepRL}
K.~He, X.~Zhang, S.~Ren, and J.~Sun, ``Deep residual learning for image
  recognition,'' \emph{2016 IEEE Conference on Computer Vision and Pattern
  Recognition (CVPR)}, pp. 770--778, 2016.

\bibitem{bert-devlin-etal-2019-bert}
\BIBentryALTinterwordspacing
J.~Devlin, M.-W. Chang, K.~Lee, and K.~Toutanova, ``{BERT}: Pre-training of
  deep bidirectional transformers for language understanding,'' in
  \emph{Proceedings of NAACL}.\hskip 1em plus 0.5em minus 0.4em\relax
  Minneapolis, Minnesota: Association for Computational Linguistics, Jun. 2019,
  pp. 4171--4186. [Online]. Available: \url{https://aclanthology.org/N19-1423}
\BIBentrySTDinterwordspacing

\bibitem{wenet-DBLP:journals/corr/abs-2102-01547}
\BIBentryALTinterwordspacing
B.~Zhang, D.~Wu, C.~Yang, X.~Chen, Z.~Peng, X.~Wang, Z.~Yao, X.~Wang, F.~Yu,
  L.~Xie, and X.~Lei, ``Wenet: Production first and production ready end-to-end
  speech recognition toolkit,'' \emph{CoRR}, vol. abs/2102.01547, 2021.
  [Online]. Available: \url{https://arxiv.org/abs/2102.01547}
\BIBentrySTDinterwordspacing

\bibitem{ctc-10.1145/1143844.1143891}
\BIBentryALTinterwordspacing
A.~Graves, S.~Fern\'{a}ndez, F.~Gomez, and J.~Schmidhuber, ``Connectionist
  temporal classification: Labelling unsegmented sequence data with recurrent
  neural networks,'' in \emph{Proceedings of ICML}, ser. ICML '06.\hskip 1em
  plus 0.5em minus 0.4em\relax New York, NY, USA: Association for Computing
  Machinery, 2006, p. 369–376. [Online]. Available:
  \url{https://doi.org/10.1145/1143844.1143891}
\BIBentrySTDinterwordspacing

\bibitem{KL-10.1214/aoms/1177729694}
\BIBentryALTinterwordspacing
S.~Kullback and R.~A. Leibler, ``{On Information and Sufficiency},'' \emph{The
  Annals of Mathematical Statistics}, vol.~22, no.~1, pp. 79 -- 86, 1951.
  [Online]. Available: \url{https://doi.org/10.1214/aoms/1177729694}
\BIBentrySTDinterwordspacing

\bibitem{Kingma2015AdamAM}
D.~P. Kingma and J.~Ba, ``Adam: A method for stochastic optimization,''
  \emph{CoRR}, vol. abs/1412.6980, 2015.

\bibitem{kudo-richardson-2018-sentencepiece}
\BIBentryALTinterwordspacing
T.~Kudo and J.~Richardson, ``{S}entence{P}iece: A simple and language
  independent subword tokenizer and detokenizer for neural text processing,''
  in \emph{Proceedings of EMNLP}.\hskip 1em plus 0.5em minus 0.4em\relax
  Brussels, Belgium: Association for Computational Linguistics, Nov. 2018, pp.
  66--71. [Online]. Available: \url{https://aclanthology.org/D18-2012}
\BIBentrySTDinterwordspacing

\bibitem{spec-aug-2019}
\BIBentryALTinterwordspacing
D.~S. Park, W.~Chan, Y.~Zhang, C.-C. Chiu, B.~Zoph, E.~D. Cubuk, and Q.~V. Le,
  ``Specaugment: A simple data augmentation method for automatic speech
  recognition,'' \emph{Interspeech 2019}, Sep 2019. [Online]. Available:
  \url{http://dx.doi.org/10.21437/Interspeech.2019-2680}
\BIBentrySTDinterwordspacing

\bibitem{cmvn-10.1016/S0167-6393(98)00033-8}
\BIBentryALTinterwordspacing
O.~Viikki and K.~Laurila, ``Cepstral domain segmental feature vector
  normalization for noise robust speech recognition,'' \emph{Speech Commun.},
  vol.~25, no. 1–3, p. 133–147, aug 1998. [Online]. Available:
  \url{https://doi.org/10.1016/S0167-6393(98)00033-8}
\BIBentrySTDinterwordspacing

\end{thebibliography}

% \begin{thebibliography}{9}
% \bibitem[1]{Davis80-COP}
%   S.\ B.\ Davis and P.\ Mermelstein,
%   ``Comparison of parametric representation for monosyllabic word recognition in continuously spoken sentences,''
%   \textit{IEEE Transactions on Acoustics, Speech and Signal Processing}, vol.~28, no.~4, pp.~357--366, 1980.
% \bibitem[2]{Rabiner89-ATO}
%   L.\ R.\ Rabiner,
%   ``A tutorial on hidden Markov models and selected applications in speech recognition,''
%   \textit{Proceedings of the IEEE}, vol.~77, no.~2, pp.~257-286, 1989.
% \bibitem[3]{Hastie09-TEO}
%   T.\ Hastie, R.\ Tibshirani, and J.\ Friedman,
%   \textit{The Elements of Statistical Learning -- Data Mining, Inference, and Prediction}.
%   New York: Springer, 2009.
% \bibitem[4]{YourName17-XXX}
%   F.\ Lastname1, F.\ Lastname2, and F.\ Lastname3,
%   ``Title of your INTERSPEECH 2022 publication,''
%   in \textit{Interspeech 2022 -- 23\textsuperscript{rd} Annual Conference of the International Speech Communication Association, September 18-22, Incheon, Korea, Proceedings, Proceedings}, 2022, pp.~100--104.
% \end{thebibliography}

\end{document}